\pdfoutput=1
\documentclass[11pt]{article}

\usepackage{NLPAICS2024}

\usepackage{times}
\usepackage{latexsym}

\usepackage[T1]{fontenc}
\usepackage[utf8]{inputenc}

\usepackage{microtype}

\usepackage{inconsolata}

\usepackage{booktabs}
\usepackage{multirow}
\usepackage[table]{xcolor}
\usepackage{algorithm}
\usepackage{algorithmicx}
\usepackage{algpseudocode}
\usepackage{amsmath}
\usepackage{amssymb}
\usepackage{threeparttable}
\usepackage{graphicx}
\usepackage{svg}
\usepackage{fancyhdr}
\usepackage{fancyvrb}

\definecolor{gray}{gray}{0.9}
\definecolor{lightgreen}{RGB}{204,255,204}
\definecolor{lightred}{RGB}{255,204,204}
\definecolor{green}{RGB}{240,255,240}
\definecolor{red}{RGB}{255,235,235}

\renewcommand{\footnotesize}{\fontsize{9}{11}\selectfont}

\title{Inducing Personality in LLM-Based Honeypot Agents: Measuring the Effect on Human-Like Agenda Generation}

\author{
\textbf{Lewis Newsham} \and \textbf{Ryan Hyland} \and \textbf{Daniel Prince} \\
School of Computing and Communications \\
Lancaster University, UK \\
\texttt{\{lewis.newsham, r.hyland, d.prince\}@lancaster.ac.uk}
}

\fancypagestyle{NLPAICS}{
  \fancyhf{}
  \fancyfoot[C]{\textit{Proceedings of the 1st International Conference on NLP \& AI for Cyber Security}, pages 48--58\\July 29--30, 2024.}

}

\begin{document}
\maketitle
\thispagestyle{NLPAICS}

\begin{abstract}

    This paper presents SANDMAN, an architecture for cyber deception that leverages Language Agents to emulate convincing human simulacra. Our `Deceptive Agents' serve as advanced cyber decoys, designed for high-fidelity engagement with attackers by extending the observation period of attack behaviours. Through experimentation, measurement, and analysis, we demonstrate how a prompt schema based on the five-factor model of personality systematically induces distinct `personalities' in Large Language Models. Our results highlight the feasibility of persona-driven Language Agents for generating diverse, realistic behaviours, ultimately improving cyber deception strategies.

\end{abstract}

\section{Introduction}\label{Sec:Introduction}

Autonomous agents are systems embedded within environments, capable of autonomous interaction to influence future conditions, driven by programmed objectives \cite{franklin1996agent, BOSSER20011002}. Historically, agent autonomy was enabled through simple heuristic policies or learned behaviours within defined constraints \cite{schulman2017proximal, mnih2015human, lillicrap2015continuous}. However, recent advances in the field of Generative Artificial Intelligence (Gen-AI) are radically transforming intelligent agent technologies. The most noteworthy and pertinent are Large Language Models (LLMs) which have demonstrated a remarkable ability to generate human-like text, answer complex questions, and perform other language-driven tasks with high accuracy \cite{floridi2020gpt, kasneci2023chatgpt}. As such, there is growing interest in applying these models as autonomous agent controllers to yield more human-like decision-making capabilities \cite{chen2019generative, shinn2024reflexion, shen2024hugginggpt}. This approach exploits an LLM's comprehensive internal model of the world, enhanced by transformer architectures that capture long-range dependencies in text \cite{vaswani2017attention}, to inform actions without domain-specific training. In parallel, researchers have extended LLMs with memory and planning functions to enhance an agents' human-like capabilities \cite{park2023generative, hong2023metagpt, qian2023communicative}, leading to the concept of Language Agents \cite{kenton2021alignment, zhou2023agents, sumers2023cognitive}.

Novel applications using autonomous agents within security-centric applications include: automating red teaming exercises \cite{happe2023getting, deng2023pentestgpt}, enhancing anomaly detection systems \cite{ott2021robust, su2024large} and, streamlining threat intelligence analysis \cite{bayer2023multi}. However, to the best of our knowledge, no research has explored their application suited for Active Cyber Defense strategies \cite{denning2014framework}, aimed at disrupting early stage cyber-adversary activities \cite{yadav2015technical}. Cyber Deception research focuses on game-theoretic techniques \cite{pawlick2019game} and deception technology \cite{spitzner2003honeypots} to deceive malicious actors via means of mimicry, camouflage, obfuscation etc. This paper introduces the concept of \textbf{Deceptive Agents} as entities employing generative models to deceive attackers with plausible (mis-)information and behaviours to disrupt attack progress. Our work presents an architecture to endow agents with the capability to accumulate, synthesise, and utilise memories facilitating the generation of contextually relevant, plausible behaviour that dynamically adjusts to experiences and environments. In summary, this paper makes the following contributions:

\begin{itemize}
    \setlength{\itemsep}{0em}
    \setlength{\parskip}{0pt}
    \setlength{\parsep}{0pt}
        \item \textit{Deceptive Agents} architecture to create plausible simulacra of human behaviour for defensive deception in digital environments;
        \item A prompting schema to control the generation of Deceptive Agent personalities;
        \item An evaluation method to demonstrate the impact of induced personality within agents.
\end{itemize}

The paper is structured as follows: Section \ref{Sec:Related Work} outlines related work, Section \ref{Sec:Architecture} presents the SANDMAN architecture to operate deceptive agents, Section \ref{Sec:Experiments} outlines experiments and analyses performed concerning the controlled induction of personas within LLMs based on the five-factor model (FFM), Section \ref{Sec:Discussion} provides a discussion of the findings, including directions for future work, and Section \ref{Sec:Conclusion} presents the conclusion.
\section{Related Work} \label{Sec:Related Work}

Prior research has explored design considerations and behaviours of autonomous agents, the utility and efficacy of LLMs in security-focused applications, and identifying existing issues within traditional defensive deception strategies. These are key domains of study to realising \textit{Deceptive Agents}. 

\paragraph{LLMs in Defensive Applications:} Gen-AI presents a series of new opportunities for cyber-security. Researchers have  explored utilising LLMs within security-focused applications, demonstrating their potential in automating and streamlining complex security processes. Notable advancements include their application to software security testing \cite{happe2023getting}, log-based analytics \cite{ma2024llmparser, setianto2021gpt}, unstructured text analysis for threat intelligence \cite{bayer2023multi}, and security-based training \cite{gundu2023chatbots}.

\paragraph{Language Agents:} An emerging class of autonomous agent leveraging LLMs as central controllers to direct actions \cite{sumers2023cognitive, hong2023metagpt, kenton2021alignment, zhou2023agents}. Research has introduced bespoke architectures and frameworks for language agents (LAs) providing varied applications across diverse environments. These include the simulation of multi-agent sandbox environments to study inter-agent behaviour \cite{park2023generative}, collaborative frameworks in software development \cite{qian2023communicative}, and the integration of agents within video games \cite{wang2023voyager}. These studies underscore the proficiency of LLMs to manage complex, autonomous agent behaviours. However, the existing literature primarily explores these agents in non-security contexts or in scenarios where the environment or application sets inherent limitations on their utility.

\paragraph{Agent Architectures:} Whilst the concept of LAs is relatively straightforward (\textit{i.e.}, using a LLM as an autonomous agent controller), achieving the intended effect (\textit{i.e.}, long-horizon task completion) is typically far more complex \cite{wang2024survey}. This has led to new frameworks to categorise existing agents and plan future developments. The Cognitive Architecture for Language Agents (CoALA), is a comprehensive approach which draws on cognitive science and symbolic AI to characterise general purpose architectures for LAs \cite{sumers2023cognitive}. CoALA organises agents along three key dimensions: their \textit{information storage} (memories); \textit{action space} (internal/external); and \textit{decision-making procedures} (interactive loop with planning and execution). The core components of the CoALA framework are provided below:

\begin{itemize}
    \setlength{\itemsep}{0em}
    \setlength{\parskip}{0pt}
    \setlength{\parsep}{0pt}
    % \footnotesize
    \item \textbf{Decision Procedure:} Engine to interconnect modular components and execute agent code
    \item \textbf{Procedural Memory:} Implicit (LLM) and explicit (programmatic) knowledge for dictating functionality and decision-making
    \item \textbf{Semantic Memory:} Agent's repository of structured knowledge about itself which evolves following interaction with the environment, enhancing its knowledge base
    \item \textbf{Episodic Memory:} Dynamic module to capture and store experiences and decisions from past interactions to inform and contextualise decisions and actions
    \item \textbf{Working Memory:} Temporarily holds and manages information (\textit{i.e.}, active knowledge) relevant to the current decision cycle
\end{itemize}

\paragraph{Gray Agent Simulation:} Effective and plausible pattern-of-life behaviour emulation within gray agents and non-playable characters (NPCs) remains an active area of research. A pertinent example in the context of this work is the GHOSTS framework \cite{updyke2018ghosts}. Agents in GHOST emulate user behaviour within digital environments to exhibit stochastic behaviour which is suited toward training and cyber-based exercises.
\section{Deceptive Agent Architecture} \label{Sec:Architecture}

In this section we provide the architecture for SANDMAN, a software platform for AI-driven autonomous agents in generating plausible behaviours within a digital environment. At its core, SANDMAN represents a novel contribution to the emerging research field concerning cognitive architectures and language agents \cite{sumers2023cognitive}. Modular and extensible by design, SANDMAN enables fine-tuning of agents to support various applications, ranging from human-like gray agent simulation for cyber-warfare exercising and defender emulation, to augmenting deception platforms to provide dynamic and plausible environments.

\begin{figure}[htb]
    \centering
    \includegraphics[width=\linewidth]{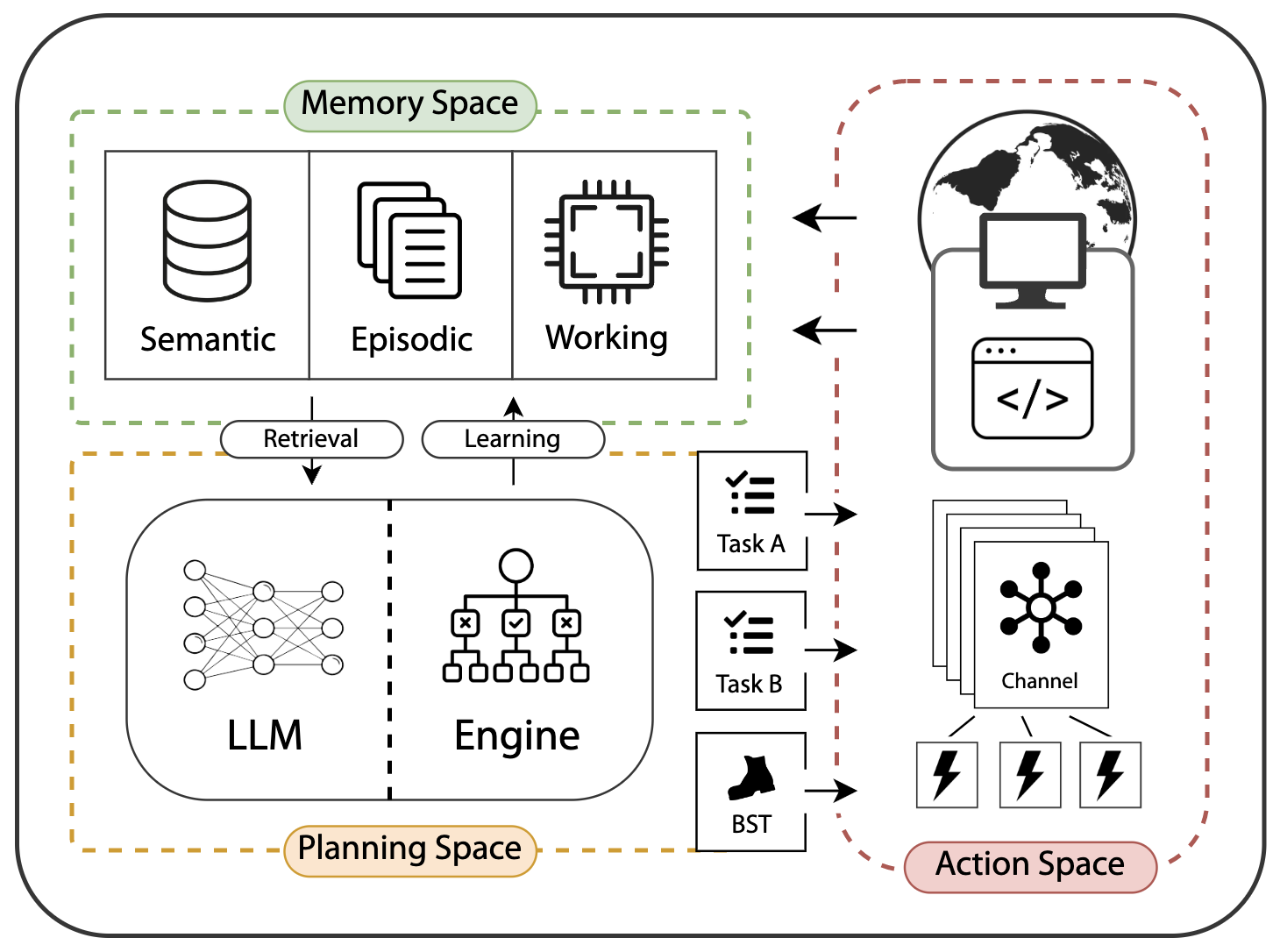}
    \captionsetup{font=small}
    \caption{Architecture for SANDMAN agents, inspired by CoALA framework \cite{sumers2023cognitive}.}
    \label{fig:agent-architecture}
\end{figure}

The goal of SANDMAN is not to interact with other humans or agents. Rather, it is intended to produce plausible simulacra representing human-like actions in digital environments that, to the observer, cannot be distinguished from human. A particular focus area SANDMAN seeks to address concerns \textit{generative deception}, a novel concept that, to the best of our knowledge, has not yet been explored in the context of autonomous agents.

\textbf{Agent Profile:} The crux of definable agent behaviour is rooted within agent profiling, a method to construct the personalities of singular agents \cite{wang2024survey}. For SANDMAN, whose purpose is to facilitate its agents in generating human-like patterns of thought and belief, by virtue of their actions, considerable emphasis is placed upon controlled personality induction. Construction of an agent's personality is discussed in Section \ref{Sec:Inducing Personality in LLMs}, whereas its induced effect is empirically evaluated and analysed in Section \ref{Sec:Experiments}.

\textbf{Decision Engine:} Central to a SANDMAN agent is its ability to decide what to do at any given time. Pivotal to task selection and execution is a decision engine: the central processing component. The Decision Engine can be considered the top-level or "main" agent program. It dynamically observes and handles all internal processes at runtime, acting as overseer; synergising various memory components with task-oriented modules whilst managing decision-making.

\textbf{Memory:} A critical pillar in LA design, serving various functions to support reasoning and learning \cite{wang2024survey}. SANDMAN uses a common memory architecture that can be used for semantic, episodic, and procedural purposes \cite{sumers2023cognitive}. In addition, the platform extensively uses 'working' memory, a generic store across all components to facilitate reflective operation, a nuanced form of reasoning and retrieval. Memory ensures agents remain on-task, contextually rich, and grounded in the environment whilst adhering to specifications, such as prompt templates (procedural) and structured profiling (semantic).

\textbf{Task List:} Represents all possible actions made available to an agent at a given point. Task categories are inspired by those in GHOSTS \cite{updyke2018ghosts}, featuring  work and non-work related tasks. Initialised by the bootstrap task (BST), the task list also embodies episodic memory--recursively queried to contextualise future actions based on previous decision-cycles. The task list is designed to shrink and grow as an agent completes tasks and as new tasks are generated, enabling dynamic and continuous behaviour. Task modules can be reflectively loaded by SANDMAN enabling easier modular development.
The Bootstrap Task is essential to the planning of the agents activities for the day. Section \ref{Sec:Experiments} explores the use of an LLM (GPT3.5-Turbo) to generate schedules from a list of available tasks which SANDMAN can load. Our LLM-based BST module is \textit{PlanScheduleTask}, which prompts the LLM with its agent profile (semantic), other forms of memory, and the available task list. The LLM will then return a list of tasks and add them to an agent's task list, with the decision engine then deciding on what task to perform next.

\textbf{Channel:} For agents' actions to manifest and become tangible to the observer, an intermediate channel module is required. Channels are situated between tasks and the environment. Their purpose is to hook an assigned task to the appropriate end-application, in essence bringing SANDMAN to 'life' by eliciting an action in the environment. The positioning of channel modules enables SANDMAN to interact with various parts of the underlying system it is interfacing with. Channels can therefore be considered abstraction layers wrapped around user applications providing a common API. For example, the WebChannel module wraps around the Firefox browser, enabling user-like interactions with the browser itself (\textit{e.g.}, typing in the address bar, scrolling the page). All these procedural actions are governed within distinct channels. The key strength of this is extensibility; channels can be added, modified, or removed depending on the intended purpose by the end-user.

\textbf{Generators:}  API calls for LLM-generated content needed to complete tasks is performed by generators. The models are prompted with an agent persona, memory and task metadata to generate the necessary content to complete a task. This content is then passed to a channel that accepts generated content as an input to use when interfacing with a program. For example, a 'write document' task will have a 'Microsoft Word' channel to interface with Microsoft Word. Content to populate the Word document will be provided by a generator with an LLM that the channel uses as an input to then type, in a simulated manner to reflect human-like type speed which may features mistakes, the generated content into the word document.
\section{Persona-based Task Planning in LLMs} \label{Sec:Experiments}

Planning modules are essential for autonomous agents, enabling structured and controlled behaviour \cite{sumers2023cognitive, wang2024survey}. As demonstrated in existing studies \cite{park2022social, hong2023metagpt, qian2023communicative}, planning heavily influences activities performed by agent(s).

In SANDMAN, the planning functionality is provided by the Bootstrap Task. Initial debugging and development used a simple rule-based approach to generate an agent's plan, validating that the execution flow aligned with the architecture's design. This approach involved appending tasks sequentially in a straightforward, deterministic manner. However, task scheduling via an LLM presents a novel and unexplored opportunity. Although LLMs have been used similarly in other contexts \cite{park2023generative}, there has been no systematic investigation into the relationship between persona generation and the resulting task outputs. Typically, the variance in outputs is either asserted or assumed without rigorous analysis. In this section, we demonstrate the structured creation and induction of personas into LLMs presents distinct effects on associated, LLM-generated schedules.

\subsection{Inducing Personality Types in LLMs} \label{Sec:Inducing Personality in LLMs}

Autonomous agents in recent studies which leverage LLMs typically perform tasks by assuming specific roles, such as coder, teacher, domain expert etc. \cite{wang2024survey}. Agent profiling is an approach to construct unique personas, either through handcrafting \cite{park2023generative}, LLM-based generation \cite{zhang2023building}, or dataset alignment \cite{argyle2023out}, to encompass definable characteristics such as name, role, occupation, and passion etc. As per the CoALA framework and prior approaches, these are stored in semantic memory and passed in at various stages within decision-cycles to contextualise internal and external action spaces, such as reasoning and retrieval, and grounding, respectively \cite{sumers2023cognitive}. For instance, \textit{"John Lin is a pharmacy shopkeeper at the Willow Market and Pharmacy who loves to help people. He is always looking for ways to make the process ..."} \cite{park2023generative}. 

%In each case, a handcrafting or LLM-based method is often sought, thereby the persona generaton process is dependent upon either internal biases of the developer or knowledge weightings in the LLM. 
The choice of information to profile the agent is largely determined by the specific application scenario(s) \cite{wang2024survey}. Therefore, if the intended purpose is to generate believable proxies of human behaviour, peronas ought to be crafted using descriptors rooted in psychology theory. The recent work of \citet{safdari2023personality, jiang2024evaluating} demonstrates that LLMs can be induced to appropriately respond to human psychometric assessment methods through crafted prompts.

The Machine Personality Inventory (MPI) by \citet{jiang2024evaluating} systematically evaluates machines' personality-like behaviours in psychometric tests against the Big-Five Personality Traits: \underline{O}penness, \underline{C}oncientiousness, \underline{E}xtraversion, \underline{A}greeableness, \underline{N}euroticism (OCEAN) \cite{costa1999five, mccrae1997personality}. The MPI adapts the International Personality Item Pool (IPIP) \cite{goldberg1999broad, goldberg2006international, johnson2014measuring} to psychometrically test LLMs akin to how psychologists evaluate humans. The MPI features 24 distinct statements pertaining to each OCEAN factor. For instance, "Love to help others" is associated with an individual high in \underline{A}greeableness. The LLM is then instructed provide an answer to this statement based on its own self-perception, ranging from "(A). Very Accurate" to "(E). Very Inaccurate". Once complete, the results are calculated and evaluated as one would with a human subject. \citet{jiang2024evaluating} demonstrated that, through crafted prompts, it is possible to induce personalities traits correlating with specifc persona prompts. However, it is noted the evaluation and results of this work primarily emphasise the effect of positive-induction only, largely discarding the effect following negative induction. Moreover, results from the experiments performed therein are not rigorously scrutinised or subjected to statistical testing to measure for significance between OCEAN scores from the experimental (LLM) and control (human) groups.

\subsubsection{Experimental Method} \label{Experimental Method}

We incorporate the MPI to verify that our choosen LLM for task planning (GPT-3.5-Turbo) exhibits similar performance to that of previous models evaluated by \citet{jiang2024evaluating}, such as BART, GPT-Neo 2.7B, Alpaca 7B etc. To that end, an adapted prompt strategy is employed in our experiments, combining what is referred to as \textbf{Naive}- and \textbf{Words}-based prompting methods \cite{jiang2024evaluating}. In the context of personality, the former involves using a standard naive natural language prompt (\textit{i.e.}, "You are extraverted"), and the latter involves prompt search (\textit{i.e.}, "outgoing, energetic, public"), one of the most effective prompting methods \cite{prasad2022grips, shin2020autoprompt}. This was done to ensure for clear causal linkage between dependent (personality trait) and independent (MPI Score) variables without introducing uncertainty via any intermediate interpretation (such as through an LLM).  The personality trait schema is therefore:

\begin{center}

\fontsize{9pt}{11pt}\selectfont
"Imagine you are a/an \( X \) person characterised by being \( Y \)", where \( X \) is the naive title of the Big-Five trait, for example \textit{Extraverted} and where \( Y \) are descriptive words assocated with the trait such as \textit{outgoing, energetic, public}. 
\end{center}

Each personality prompt is passed through the MPI 5 times, with the averages across all the responses recorded. A baseline, control data set is produced by prompting the LLM without a personality trait statement in the prompt. The LLM has \texttt{Temperature (0.7)} for all trials. As per \citet{jiang2024evaluating}, we calculate the mean ($\mu$) and standard deviation ($\sigma$) of the personality items, but we use two-sampled t-test for significance (p $\leq$ 0.05).

Table \ref{tab:gpt_3.5_0125_OCEAN_results} presents the computed MPI scores across experimental conditions, highlighting the efficacy of controlled personality induction within an LLM. Each induced OCEAN trait (+/-) yielded a statistically significant score for the targeted trait when compared against the control condition (Neutral), thereby confirming the effectiveness of our prompting schema and method of induction on the opted model (GPT-3.5-Turbo). A bleed-through effect is also observed, indicating cross-trait influences.

While the personality trait schema is appropriate for the experiments discussed later in the paper, the evaluation method described could also be used to refine trait schemas to achieve specific outcomes. For example, word-based selection can be adjusted to either reduce or enhance bleed-through, or to modulate the t-score to either strengthen or weaken deviations from the baseline while maintaining statistical significance.

\begin{table}[H]
\centering
\small
\captionsetup{font=small}
\vspace{-2pt}
% \caption{Single-factor personality analysis on opted LLM (GPT-3.5-Turbo). Highlighted cells in gray denote statistical significance at $p$ $\leq$ 0.05 level. \footnotesize\textsuperscript{1}Control group.}
% \label{tab:gpt_3.5_0125_OCEAN_results}
% \renewcommand{\arraystretch}{0.7} % Default: 1

\begin{tabular}{@{}c l @{\hspace{7pt}}l @{\hspace{9pt}} c @{\hspace{9pt}} c @{\hspace{9pt}} c @{\hspace{9pt}} c @{\hspace{9pt}} c @{\hspace{9pt}} c @{\hspace{9pt}} c @{\hspace{9pt}} c @{\hspace{9pt}} c @{\hspace{9pt}} c @{\hspace{9pt}} c @{\hspace{9pt}} c @{\hspace{9pt}} c @{\hspace{9pt}} c @{\hspace{9pt}} c c@{}}
%\toprule

\toprule
& & \textit{Dir} & \textbf{O} & \textbf{C} & \textbf{E} & \textbf{A} & \textbf{N} \\

% & & & Score & $t$-val & $p$-val & Score & $t$-val & $p$-val & Score & $t$-val & $p$-val & Score & $t$-val & $p$-val & Score & $t$-val & $p$-val & \\
\midrule

& \multirow{2}{*}{\textbf{O}} & Pos & \cellcolor{lightgreen}\textbf{4.30$^*$} & 3.72 & 4.02 & \cellcolor{gray}\textbf{4.23} & \cellcolor{gray}\textbf{2.29}  \\
& & Neg & \cellcolor{lightred}\textbf{2.07} & 4.10 & \cellcolor{gray}\textbf{2.10$^*$} & 3.49 & 2.64 \\
\midrule

& \multirow{2}{*}{\textbf{C}} & Pos & 3.36 & \cellcolor{lightgreen}\textbf{4.83$^*$} & 3.25 & \cellcolor{gray}\textbf{4.28} & \cellcolor{gray}\textbf{1.96} \\
& & Neg & 4.00 & \cellcolor{lightred}\textbf{2.02$^*$} & \cellcolor{gray}\textbf{2.35} & 3.66 & 3.64 \\
\midrule

& \multirow{2}{*}{\textbf{E}} & Pos & 3.66 & 3.64 & \cellcolor{lightgreen}\textbf{4.67$^*$} & 3.98 & \cellcolor{gray}\textbf{2.36} \\
& & Neg & 3.17 & 2.98 & \cellcolor{lightred}\textbf{1.46$^*$} & 3.69 & 3.48 \\
\midrule

& \multirow{2}{*}{\textbf{A}} & Pos & 3.57 & 3.94 & 3.31 & \cellcolor{lightgreen}\textbf{4.72$^*$} & 2.40 \\
& & Neg & 2.73 & 2.65 & 2.92 & \cellcolor{lightred}\textbf{2.12$^*$} & 3.40 \\
\midrule

& \multirow{2}{*}{\textbf{N}} & Pos & 3.54 & \cellcolor{gray}\textbf{2.55} & \cellcolor{gray}\textbf{2.60} & 3.82 & \cellcolor{lightgreen}\textbf{4.50$^*$} \\
& & Neg & \cellcolor{gray}\textbf{3.44} & 4.22 & 3.35 & \cellcolor{gray}\textbf{4.27} & \cellcolor{lightred}\textbf{1.32$^*$} \\
\midrule

& \textbf{B}\textsuperscript{1} & N/A & \textbf{3.33} & \textbf{3.55} & \textbf{3.65} & \textbf{3.39} & \textbf{3.04} \\

\bottomrule
\end{tabular}
\caption{Single-factor personality analysis on opted LLM (GPT-3.5-Turbo). Highlighted cells in gray denote statistical significance at $p$ $\leq$ 0.05 level. \footnotesize\textsuperscript{1} Control group.}
\label{tab:gpt_3.5_0125_OCEAN_results}
\end{table}
\vspace{-15pt}

\begin{table*}[htb]
\captionsetup{font=small}
\scriptsize
% \caption{Comparison of treatment groups (Sys, Rand, Sys \& Rand) for Task Duration and Frequency.  Values are Means ($\mu$) and Std. Dev. ($\sigma$) in parentheses. Highlighted cells in gray denote statistically significant deviations ($p$ $\leq$ 0.05) from either the corresponding task Duration or Frequency within the control (Baseline) condition.}
% \label{tab:schedule_compare_duration_count}
    \centering
    \setlength{\tabcolsep}{8pt}
        \begin{tabular}{@{}r c c c c c c c c@{}}
            \toprule

            & 
            \multicolumn{2}{c}{\centering \textbf{Baseline}} & 
            \multicolumn{2}{c}{\centering \textbf{Sys}} & 
            \multicolumn{2}{c}{\centering \textbf{Rand}} & 
            \multicolumn{2}{c}{\centering \textbf{Sys} \textbf{\&} \textbf{Rand}} \\
            
            \cmidrule(l){2-3}
            \cmidrule(l){4-5} 
            \cmidrule(l){6-7} 
            \cmidrule(l){8-9}
            
            \textbf{Task} & 
            Duration & Frequency &
            Duration & Frequency &
            Duration & Frequency &
            Duration & Frequency \\
            
%            \cmidrule(l){2}\cmidrule(l){3} 
%            \cmidrule(l){4}\cmidrule(l){5}
%            \cmidrule(l){6} \cmidrule(l){7} 
%            \cmidrule(l){8} \cmidrule(l){9}
%            
%            Task & 
%            $\mu$ ($\sigma$) & $\mu$ ($\sigma$) &  
%            $\mu$ ($\sigma$) & $\mu$ ($\sigma$) &  
%            $\mu$ ($\sigma$) & $\mu$ ($\sigma$) &  
%            $\mu$ ($\sigma$) & $\mu$ ($\sigma$)\\
            \midrule

Call & 59.51 (5.06) & 0.98 (0.15) & \cellcolor{gray}\textbf{55.08 (15.24)} & 0.97 (0.18) & \cellcolor{gray}\textbf{53.63 (12.49)} & 0.72 (0.45) & \cellcolor{gray}\textbf{46.44 (14.50)} & 0.92 (0.29) \\

Coffee & 56.07 (10.22) & 0.86 (0.34) & \cellcolor{gray}\textbf{31.09 (7.74)} & 0.88 (0.32) & \cellcolor{gray}\textbf{44.31 (18.37)} & 0.70 (0.46) & \cellcolor{gray}\textbf{31.35 (12.65)} & 0.89 (0.32) \\

Creative & 61.98 (7.93) & 1.00 (0.00) & \cellcolor{gray}\textbf{73.19 (14.46)} & 1.00 (0.08) & 61.52 (9.78) & 0.90 (0.34) & 62.27 (14.19) & \cellcolor{gray}\textbf{1.00 (0.24)} \\

Email & 57.23 (8.83) & 1.01 (0.13) & \cellcolor{gray}\textbf{36.78 (12.15)} & \cellcolor{gray}\textbf{1.16 (0.37)} & \cellcolor{gray}\textbf{53.44 (12.78)} & 0.77 (0.44) & \cellcolor{gray}\textbf{43.42 (13.79)} & \cellcolor{gray}\textbf{0.97 (0.32)} \\

Exercise & 59.35 (5.89) & 0.93 (0.26) & \cellcolor{gray}\textbf{52.82 (12.05)} & 0.91 (0.29) & \cellcolor{gray}\textbf{58.20 (9.25)} & 0.76 (0.43) & \cellcolor{gray}\textbf{55.78 (11.95)} & 0.93 (0.26) \\

Reading & 57.53 (8.26) & 0.93 (0.26) & \cellcolor{gray}\textbf{42.65 (12.33)} & 0.94 (0.24) & \cellcolor{gray}\textbf{54.81 (11.58)} & 0.68 (0.48) & \cellcolor{gray}\textbf{47.11 (13.27)} & 0.94 (0.25) \\

Lunch & 60.18 (3.00) & 1.00 (0.00) & \cellcolor{gray}\textbf{63.95 (11.09)} & 1.00 (0.04) & 60.06 (3.57) & 1.00 (0.04) & 60.00 (9.23) & 1.00 (0.04) \\

Meeting & 61.86 (7.49) & 1.00 (0.00) & \cellcolor{gray}\textbf{69.37 (15.78)} & 1.00 (0.06) & \cellcolor{gray}\textbf{60.17 (9.32)} & 0.91 (0.30) & 60.95 (14.35) & 0.96 (0.24) \\

Break & 55.23 (10.99) & 0.94 (0.25) & \cellcolor{gray}\textbf{37.57 (13.08)} & \cellcolor{gray}\textbf{1.01 (0.28)} & \cellcolor{gray}\textbf{49.92 (14.57)} & 0.75 (0.47) & \cellcolor{gray}\textbf{36.69 (12.95)} & \cellcolor{gray}\textbf{0.99 (0.34)} \\

Personal & 57.48 (8.85) & 0.96 (0.21) & \cellcolor{gray}\textbf{44.25 (14.44)} & \cellcolor{gray}\textbf{0.98 (0.23)} & 56.92 (11.44) & 0.88 (0.37) & \cellcolor{gray}\textbf{48.39 (14.26)} & \cellcolor{gray}\textbf{1.06 (0.34)} \\

Plan & 59.75 (3.83) & 0.98 (0.13) & 59.57 (13.68) & 0.98 (0.17) & \cellcolor{gray}\textbf{57.55 (9.25)} & 0.87 (0.35) & \cellcolor{gray}\textbf{54.09 (13.84)} & \cellcolor{gray}\textbf{1.00 (0.19)} \\

Reflect & 53.16 (13.91) & 0.95 (0.23) & \cellcolor{gray}\textbf{40.32 (12.69)} & 0.99 (0.20) & 54.45 (11.80) & \cellcolor{gray}\textbf{0.98 (0.26)} & \cellcolor{gray}\textbf{46.48 (14.09)} & \cellcolor{gray}\textbf{1.05 (0.30)} \\

Research & 59.24 (5.88) & 0.88 (0.32) & 58.14 (13.91) & 0.98 (0.14) & 59.57 (8.60) & 0.93 (0.29) & 60.31 (14.28) & 1.00 (0.13) \\

Media & 57.35 (9.77) & 0.96 (0.21) & \cellcolor{gray}\textbf{42.29 (13.26)} & 0.93 (0.25) & \cellcolor{gray}\textbf{53.55 (13.09)} & \cellcolor{gray}\textbf{0.75 (0.46)} & \cellcolor{gray}\textbf{42.64 (13.30)} & \cellcolor{gray}\textbf{0.94 (0.29)} \\

Collab. & 61.27 (7.19) & 0.96 (0.20) & \cellcolor{gray}\textbf{63.33 (13.11)} & 0.99 (0.12) & 62.32 (10.14) & 0.97 (0.17) & \cellcolor{gray}\textbf{66.25 (15.80)} & \cellcolor{gray}\textbf{1.01 (0.13)} \\

Work & 122.84 (32.84) & 1.01 (0.13) & \cellcolor{gray}\textbf{80.76 (14.54)} & \cellcolor{gray}\textbf{1.16 (0.37)} & \cellcolor{gray}\textbf{68.97 (19.24)} & 0.93 (0.31) & \cellcolor{gray}\textbf{73.17 (15.93)} & \cellcolor{gray}\textbf{1.06 (0.36)} \\

            \midrule
Reject & ... & ... & 14 & 4 & 10 & 2 & 12 & 9 \\
            
            \bottomrule
        \end{tabular}
        \caption{Comparison of treatment groups (Sys, Rand, Sys \& Rand) for task duration and frequency.  Values are means ($\mu$) and std. dev. ($\sigma$) in parentheses. Highlighted cells in gray denote statistically significant deviations ($p$ $\leq$ 0.05) from either the corresponding task duration or frequency within the control (baseline) condition.}
        \label{tab:schedule_compare_duration_count}
        \vspace{-5pt}
\end{table*}

\subsection{Persona-based Task Selection}

Given the capability to instil personality traits in LLMs, it is crucial for SANDMAN to show that these traits lead to appropriate variations in schedule generation. We measure variation via two dependent variables: (1) frequency of task occurrence in a schedule, and (2) duration of tasks within schedules, analysed on a per-task basis. To assess the impact of the independent variables (the OCEAN traits), it was necessary to establish and evaluate a suitable baseline or neutral sample. For comprehensive analysis, we generate 500 schedules using the opted LLM with \texttt{Temperature=0.7}. The fundamental procedure involves passing a list of tasks to the Boot Strap Process, which then generates the schedule for the agent to execute.

\subsubsection{Neutral Task Behaviour} \label{Neutral Task Behaviour}

In psychometric testing, establishing a baseline is essential for comparing variations across different persona types. This is equally important here, enabling observations regarding whether a given induced persona fails. Initial trials revealed a strong correlation between the order of tasks in a list and their subsequent positions in the schedule. To address this, two interventions were tested: introducing a system message and uniformly randomising the order of tasks presented in the list:

\begin{table}[H]
    \footnotesize
    \captionsetup{font=small}
    \vspace{3pt}
    % \caption{Schedule Positions. Values are Means ($\mu$) with StdDev ($\sigma$) in parentheses, and correlation coefficient ($\rho$).}
    % \label{tab:schedule_compare_position}
    \centering
    \setlength{\tabcolsep}{3.5pt}
    \begin{tabular}{@{}r cccc@{}}
        &
        \multicolumn{2}{c}{\centering \textbf{Rand}} &
        \multicolumn{2}{c}{\centering \textbf{Sys \& Rand}}\\
        \cmidrule(l){2-3}
        \cmidrule(l){4-5}
        \textbf{Task} & \textit{$\mu$ ($\sigma$)} & \textit{$\rho$} & \textit{$\mu$ ($\sigma$)} & \textit{$\rho$} \\
        \midrule
Call & 8.56 (4.78)&0.75& 8.77 (4.42)&0.63\\
Coffee & 7.51 (5.52)&0.68& 7.87 (5.92)&0.54\\
Creative & 6.42 (3.87)&0.68& 7.31 (3.86)&0.5\\
Email & 7.39 (5.29)&0.73& 7.35 (5.69)&0.5\\
Exercise & 6.84 (4.04)&0.82& 7.82 (3.5)&0.71\\
Reading & 9.97 (3.01)&0.49& 11.24 (2.24)&0.39\\
Lunch & 4.06 (1.92)&0.3& 4.12 (1.22)&0.24\\
Meeting & 4.07 (3.92)&0.64& 4.63 (4.07)&0.54\\
Break & 9.77 (3.34)&0.43& 11.34 (3.62)&0.34\\
Personal & 8.92 (3.34)&0.55& 10.44 (3.97)&0.29\\
Plan & 6.56 (4.17)&0.8& 7.0 (4.6)&0.56\\
Reflect & 7.37 (3.69)&0.62& 8.89 (4.19)&0.43\\
Research & 5.2 (3.76)&0.65& 5.62 (3.47)&0.63\\
Media & 8.66 (3.76)&0.67& 10.04 (3.41)&0.53\\
Collab. & 4.14 (3.38)&0.68& 4.11 (3.26)&0.52\\
Work & 3.31 (4.25)&0.47& 3.96 (4.93)&0.4\\
    \bottomrule
    \end{tabular}
        \caption{Schedule positions. Values are means ($\mu$) with std. dev. ($\sigma$) in parentheses, and correlation coefficient ($\rho$).}
    \label{tab:schedule_compare_position}
        \vspace{-5pt}
\end{table}

The \textit{Effect on Position} of tasks in schedules from the use of the system message alone was not significant—there was a high correlation between the task list and schedule position—with the variance in position being minimally affected. Table \ref{tab:schedule_compare_position} shows the results of randomisation (Rand) and randomisation with a system message (Sys \& Rand). Given a uniformly randomised task list in the prompt across the 500 samples, we observe variability in the position of the tasks with greater variance in many of those positions. The introduction of the system message (Sys) has the effect of weakening the correlation (a reduction in the coefficient) across all tasks. In many cases, it also reduces the positional variance. Note all correlations are statistically significant at the \(p\leq0.05\) threshold.

The \textit{Effect on Task Frequency and Duration} is presented in Table \ref{tab:schedule_compare_duration_count}. The results show the impact of the introduction of both task list order randomisation and the use of a system message. Both independent variables significantly impact the duration of tasks, notably increasing variance. However, independently, there is minimal impact on the number of task populations regarding task occurrence frequency. The combination of randomisation and a system message has a broader impact on the dependent variables.

Taken together, these results indicate that the combination of a system message and randomisation produces the optimum variation across the tasks, meeting the goals of producing a baseline dataset for further persona experiments.

\begin{table*}[htbp]
  \centering
  \scriptsize
    \captionsetup{font=small}
  % \caption{Individual Task Durations (Minutes) per OCEAN (+/-) condition with sample size $n = 500$. Values are Mean ($\mu$) with Std. Dev. ($\sigma$) in parentheses. Highlighted cells in gray denote statistically significant deviations ($p$ $\leq$ 0.05) from the corresponding task duration within the control (Neutral) condition.}
  % \label{tab:t-test_tasks_duration}

  \setlength{\tabcolsep}{3.5pt} % default value is 6pt
  
  \begin{tabular}{lccccccccccc}
    \toprule
    \multirow{2}{*}{\textbf{Task}} & \multicolumn{1}{c}{\textbf{Neutral}}

    & \multicolumn{1}{c}{\textbf{O (+)}} & \multicolumn{1}{c}{\textbf{O (--)}}
    & \multicolumn{1}{c}{\textbf{C (+)}} & \multicolumn{1}{c}{\textbf{C (--)}}
    & \multicolumn{1}{c}{\textbf{E (+)}} & \multicolumn{1}{c}{\textbf{E (--)}}
    & \multicolumn{1}{c}{\textbf{A (+)}} & \multicolumn{1}{c}{\textbf{A (--)}}
    & \multicolumn{1}{c}{\textbf{N (+)}} & \multicolumn{1}{c}{\textbf{N (--)}} \\
    
    % & \multicolumn{2}{c}{\textbf{O ($\mu$|$\sigma$)}} 
    % & \multicolumn{2}{c}{\textbf{C ($\mu$|$\sigma$)}} 
    % & \multicolumn{2}{c}{\textbf{E ($\mu$|$\sigma$)}} 
    % & \multicolumn{2}{c}{\textbf{A ($\mu$|$\sigma$)}} 
    % & \multicolumn{2}{c}{\textbf{N ($\mu$|$\sigma$)}} \\

    \cmidrule(lr){2-2} \cmidrule(lr){3-4} \cmidrule(lr){5-6} \cmidrule(lr){7-8} \cmidrule(lr){9-10} \cmidrule(lr){11-12}
    %& $\mu$ ($\sigma$) & \multicolumn{1}{c}{$\mu$ ($\sigma$)} & \multicolumn{1}{c}{$\mu$ ($\sigma$)} & \multicolumn{1}{c}{$\mu$ ($\sigma$)} & \multicolumn{1}{c}{$\mu$ ($\sigma$)} & \multicolumn{1}{c}{$\mu$ ($\sigma$)} & \multicolumn{1}{c}{$\mu$ ($\sigma$)} & \multicolumn{1}{c}{$\mu$ ($\sigma$)} & \multicolumn{1}{c}{$\mu$ ($\sigma$)} & \multicolumn{1}{c}{$\mu$ ($\sigma$)} & \multicolumn{1}{c}{$\mu$ ($\sigma$)} \\
    %\midrule

Call & 51.6 (19.7) & 50.3 (18.3) & \cellcolor{gray}\textbf{48.4 (19.6)} & \cellcolor{gray}\textbf{46.5 (19.6)} & 51.2 (17.3) & \cellcolor{gray}\textbf{55.3 (19.7)} & \cellcolor{gray}\textbf{45.3 (16.2)} & \cellcolor{gray}\textbf{48.6 (18.5)} & \cellcolor{gray}\textbf{62.2 (22.4)} & 51.0 (18.4) & 49.0 (18.9) \\
Coffee & 40.9 (17.3) & \cellcolor{gray}\textbf{37.5 (15.6)} & \cellcolor{gray}\textbf{38.1 (17.8)} & \cellcolor{gray}\textbf{32.1 (12.7)} & \cellcolor{gray}\textbf{49.6 (19.3)} & \cellcolor{gray}\textbf{36.9 (13.9)} & 43.0 (20.5) & \cellcolor{gray}\textbf{34.9 (14.4)} & \cellcolor{gray}\textbf{44.4 (21.3)} & 43.0 (19.3) & 37.5 (40.4) \\
Creative & 62.0 (19.9) & \cellcolor{gray}\textbf{66.5 (16.5)} & 62.6 (18.5) & \cellcolor{gray}\textbf{72.6 (20.0)} & 61.4 (21.0) & \cellcolor{gray}\textbf{67.5 (16.8)} & \cellcolor{gray}\textbf{69.7 (19.6)} & \cellcolor{gray}\textbf{65.6 (18.8)} & \cellcolor{gray}\textbf{71.8 (19.6)} & 63.0 (17.5) & \cellcolor{gray}\textbf{69.0 (17.8)} \\
Exercise & 57.1 (17.1) & 59.0 (14.6) & \cellcolor{gray}\textbf{51.1 (13.1)} & \cellcolor{gray}\textbf{59.1 (13.1)} & 57.1 (18.6) & \cellcolor{gray}\textbf{62.5 (13.8)} & \cellcolor{gray}\textbf{54.4 (14.5)} & 57.6 (12.7) & \cellcolor{gray}\textbf{64.3 (19.0)} & 57.2 (16.5) & \cellcolor{gray}\textbf{60.9 (13.5)} \\
Reading & 51.9 (18.6) & \cellcolor{gray}\textbf{54.4 (17.9)} & 50.4 (13.0) & \cellcolor{gray}\textbf{47.4 (17.0)} & \cellcolor{gray}\textbf{55.1 (17.3)} & \cellcolor{gray}\textbf{54.5 (16.7)} & \cellcolor{gray}\textbf{62.1 (17.3)} & 53.1 (37.6) & 51.4 (15.4) & 52.2 (16.5) & 52.9 (16.4) \\
Lunch & 65.1 (19.8) & 63.1 (15.5) & 65.6 (20.1) & 64.5 (19.5) & \cellcolor{gray}\textbf{72.2 (26.8)} & 65.0 (14.9) & 66.2 (18.4) & 65.1 (18.1) & \cellcolor{gray}\textbf{72.6 (23.8)} & 65.3 (21.5) & 63.0 (14.2) \\
Meeting & 59.0 (17.8) & \cellcolor{gray}\textbf{63.8 (16.1)} & \cellcolor{gray}\textbf{69.8 (23.0)} & \cellcolor{gray}\textbf{69.5 (17.7)} & 60.1 (20.2) & \cellcolor{gray}\textbf{68.0 (16.5)} & \cellcolor{gray}\textbf{55.8 (16.1)} & \cellcolor{gray}\textbf{65.4 (17.1)} & \cellcolor{gray}\textbf{72.0 (20.5)} & \cellcolor{gray}\textbf{63.5 (17.4)} & \cellcolor{gray}\textbf{66.8 (18.6)} \\
Break & 45.0 (18.7) & 45.8 (41.4) & 43.2 (16.4) & \cellcolor{gray}\textbf{41.1 (17.1)} & \cellcolor{gray}\textbf{52.1 (20.8)} & 46.0 (18.4) & 47.3 (20.9) & \cellcolor{gray}\textbf{41.7 (17.2)} & \cellcolor{gray}\textbf{52.2 (21.2)} & \cellcolor{gray}\textbf{47.5 (18.7)} & 43.0 (17.5) \\
Personal & 51.1 (19.5) & 48.9 (16.9) & 50.0 (16.9) & \cellcolor{gray}\textbf{46.7 (18.0)} & \cellcolor{gray}\textbf{54.5 (19.9)} & 51.9 (18.6) & 51.2 (23.0) & 49.2 (20.1) & \cellcolor{gray}\textbf{55.0 (20.1)} & 49.9 (20.6) & \cellcolor{gray}\textbf{48.5 (19.4)} \\
Plan & 55.5 (18.9) & \cellcolor{gray}\textbf{58.3 (17.6)} & \cellcolor{gray}\textbf{60.1 (20.2)} & \cellcolor{gray}\textbf{50.1 (20.4)} & 53.4 (15.5) & 56.9 (18.8) & \cellcolor{gray}\textbf{60.7 (19.6)} & \cellcolor{gray}\textbf{57.8 (17.7)} & \cellcolor{gray}\textbf{63.5 (21.2)} & 56.2 (17.3) & 56.1 (17.9) \\
Reflect & 51.1 (19.0) & \cellcolor{gray}\textbf{48.4 (17.3)} & 51.8 (19.4) & \cellcolor{gray}\textbf{48.5 (20.9)} & 52.4 (18.0) & 52.9 (19.4) & 52.5 (21.8) & \cellcolor{gray}\textbf{46.9 (20.0)} & \cellcolor{gray}\textbf{53.7 (19.6)} & 52.0 (19.8) & \cellcolor{gray}\textbf{47.7 (18.8)} \\
Research & 59.5 (19.8) & \cellcolor{gray}\textbf{62.7 (16.0)} & \cellcolor{gray}\textbf{71.8 (24.7)} & \cellcolor{gray}\textbf{71.1 (21.1)} & 57.4 (18.8) & \cellcolor{gray}\textbf{63.9 (19.9)} & \cellcolor{gray}\textbf{67.3 (20.5)} & \cellcolor{gray}\textbf{62.8 (20.0)} & \cellcolor{gray}\textbf{71.4 (22.8)} & \cellcolor{gray}\textbf{63.0 (21.0)} & \cellcolor{gray}\textbf{65.0 (19.2)} \\
Media & 48.3 (15.6) & \cellcolor{gray}\textbf{52.5 (19.6)} & \cellcolor{gray}\textbf{43.3 (13.7)} & \cellcolor{gray}\textbf{44.2 (17.9)} & \cellcolor{gray}\textbf{51.1 (18.0)} & \cellcolor{gray}\textbf{57.0 (16.7)} & \cellcolor{gray}\textbf{50.9 (18.2)} & 49.2 (20.4) & \cellcolor{gray}\textbf{52.4 (16.9)} & \cellcolor{gray}\textbf{51.6 (17.4)} & 47.4 (18.1) \\
Collab. & 62.5 (19.5) & 62.5 (15.7) & \cellcolor{gray}\textbf{69.5 (21.9)} & \cellcolor{gray}\textbf{70.3 (17.6)} & 60.2 (21.7) & \cellcolor{gray}\textbf{67.8 (16.6)} & 61.9 (18.5) & \cellcolor{gray}\textbf{66.5 (19.4)} & \cellcolor{gray}\textbf{74.5 (23.5)} & 64.7 (20.9) & \cellcolor{gray}\textbf{67.5 (16.6)} \\
Work & 63.9 (19.2) & \cellcolor{gray}\textbf{69.6 (20.4)} & \cellcolor{gray}\textbf{82.3 (25.2)} & \cellcolor{gray}\textbf{85.1 (19.0)} & 66.3 (23.3) & \cellcolor{gray}\textbf{74.2 (18.8)} & \cellcolor{gray}\textbf{78.3 (21.5)} & \cellcolor{gray}\textbf{74.1 (19.0)} & \cellcolor{gray}\textbf{78.8 (20.7)} & \cellcolor{gray}\textbf{75.0 (24.6)} & \cellcolor{gray}\textbf{77.3 (18.6)} \\

\midrule

% \multirow{2}{*}{\textbf{Null}} & Accept & 9 & 6 & 2 & 8 & 6 & 8 & 11 & 2 & 11 & 7 \\
% & Reject & 6 & 9 & 13 & 7 & 9 & 7 & 4 & 13 & 4 & 8 \\

Reject & ... & 9 & 9 & 14 & 6 & 10 & 9 & 10 & 14 & 5 & 8 \\

\bottomrule

  \end{tabular}
    \caption{Individual task durations (minutes) per OCEAN (+/-) condition with sample size $n = 500$. Values are mean ($\mu$) with std. dev. ($\sigma$) in parentheses. Highlighted cells in gray denote statistically significant deviations ($p$ $\leq$ 0.05) from the corresponding task duration within the control (Neutral) condition.}
  \label{tab:t-test_tasks_duration}
  \vspace{-5pt}
\end{table*}

\subsubsection{Induced Personality Experiments} \label{Induced Personality Experiments}

Given a suitable baseline set, we can explore the impact of induced personalities in schedule creation. Our approach extends the prompt schema to include personality trait statements. We use both positive and negative personality statements as independent variables and examine their impact on task frequency and duration. Additionally, we apply a probabilistic algorithm to compute the \textit{expected schedule} for each condition by calculating and returning the most frequent task in a given schedule slot (sequence). The expected schedule for each condition is provided in Table \ref{tab:Expected_Schedule_Compact}.

After generation, validation, and processing of the experimental and control group(s), statistical tests were performed on the metrics of \textbf{task duration} and \textbf{task frequency}. For task durations, two-sample t-tests were performed to identify statistically significant population differences at the $p$ $\leq$ 0.05 level. This analysis is given in Table \ref{tab:t-test_tasks_duration}. As frequencies of task occurrences is a form of discrete data, the Chi-square test of independence was employed. Results are displayed in Table \ref{tab:chi-square_task_frequency}.

\begin{table}[H]
\centering
\scriptsize
\captionsetup{font=small}
% \vspace{-5pt}
% \caption{Calculated Expected Schedule per OCEAN (+/-) sample. Key: Call (Cal.), Coffee (Cof.), Creative (Cre.), Exercise (Exe.), Reading (Rea.), Lunch (Lun.), Meeting (Mee.), Break (Bre.), Personal Time (PT), Plan (Pla.), Reflect (Ref.), Research (Res.), Media (Med.), Teamwork (Tea.), Work (Wrk.)}

% \caption{Calculated Expected Schedule per OCEAN (+/-) condition. $n$ = sequence slot. \footnote{}Task Abbreviation Keys.}
% \label{tab:expected_schedule}

\setlength{\tabcolsep}{3pt} % smaller column spacing for clarity

\begin{tabular}{clllllllllllll}
\toprule
\textbf{$n$} & \textbf{O +} & \textbf{O -} & \textbf{C +} & \textbf{C -} & \textbf{E +} & \textbf{E -} & \textbf{A +} & \textbf{A -} & \textbf{ N+} & \textbf{N -} \\
\midrule
1 & Wrk. & Cof. & Wrk. & Pla. & Cof. & Cof. & PT & Ref. & Wrk. & PT \\
2 & Wrk. & Cre. & Wrk. & Wrk. & Med. & Tea. & Wrk. & Wrk. & Wrk. & Wrk. \\
3 & Tea. & Res. & Mee. & Tea. & Wrk. & Tea. & Cof. & Tea. & Mee. & Cof. \\
4 & Lun. & Tea. & Lun. & Mee. & Lun. & Lun. & Lun. & Tea. & Lun. & Lun. \\
5 & Lun. & Lun. & Lun. & Lun. & Lun. & Lun. & Lun. & Lun. & Lun. & Lun. \\
6 & Lun. & Lun. & Lun. & Lun. & Lun. & Lun. & Lun. & Lun. & Res. & Lun. \\
7 & Res. & Pla. & Res. & Res. & Exe. & Exe. & Bre. & Res. & Cal. & Res. \\
8 & Cre. & Exe. & Ref. & Cre. & Tea. & Exe. & Bre. & Exe. & Exe. & Res. \\
9 & Pla. & Med. & Ref. & Ref. & Exe. & Res. & Bre. & Exe. & Cre. & Exe. \\
10 & Pla. & Med. & Ref. & Ref. & Cre. & Ref. & PT & Rea. & Exe. & Bre. \\
11 & PT & Rea. & Exe. & Exe. & Rea. & Rea. & End. & Rea. & Med. & Exe. \\
12 & Med. & Rea. & Rea. & Med. & Pla. & Rea. & Med. & Rea. & Med. & Rea. \\
13 & Rea. & Rea. & Rea. & Rea. & Pla. & Rea. & Med. & Med. & Rea. & Rea. \\
14 & Rea. & Cal. & Rea. & Rea. & Pla. & Rea. & Med. & Cal. & Rea. & Rea. \\
15 & Cof. & Ema. & Med. & Cal. & Pla. & Ema. & Cal. & Cal. & Rea. & Rea. \\
16 & Bre. & PT & Bre. & Bre. & End. & Bre. & Mee. & Bre. & Bre. & Cal. \\
17 & End. & End. & End. & End. & End. & End. & End. & End. & End. & End. \\
\bottomrule
\end{tabular}
\caption{Calculated expected schedule per OCEAN (+/-) condition. $n$ = sequence slot. \footnote{}Task abbreviation keys.}
\label{tab:Expected_Schedule_Compact}
\vspace{-5pt}
\end{table}
\footnotetext{Key: Call (Cal.), Coffee (Cof.), Creative (Cre.), Exercise (Exe.), Reading (Rea.), Lunch (Lun.), Meeting (Mee.), Break (Bre.), Personal Time (PT), Plan (Pla.), Reflect (Ref.), Research (Res.), Media (Med.), Teamwork (Tea.), Work (Wrk.)}

\begin{table*}[htbp]
  \centering
  \scriptsize
\captionsetup{font=small}
% \caption{Individual task frequency per OCEAN (+/-) condition with sample size $n = 500$. Values are mean ($\mu$) with std. dev. ($\sigma$) in parentheses. Highlighted cells in gray denote statistically significant deviations ($p$ $\leq$ 0.05) from the corresponding task frequency within the control (Neutral) condition.}
%   \label{tab:chi-square_task_frequency}

  \setlength{\tabcolsep}{3.5pt} % default value is 6pt
  
  \begin{tabular}{lccccccccccc}
    \toprule
    \multirow{2}{*}{\textbf{Task}} & \multicolumn{1}{c}{\textbf{Neutral}}

    & \multicolumn{1}{c}{\textbf{O (+)}} & \multicolumn{1}{c}{\textbf{O (-)}}
    & \multicolumn{1}{c}{\textbf{C (+)}} & \multicolumn{1}{c}{\textbf{C (-)}}
    & \multicolumn{1}{c}{\textbf{E (+)}} & \multicolumn{1}{c}{\textbf{E (-)}}
    & \multicolumn{1}{c}{\textbf{A (+)}} & \multicolumn{1}{c}{\textbf{A (-)}}
    & \multicolumn{1}{c}{\textbf{N (+)}} & \multicolumn{1}{c}{\textbf{N (-)}} \\
    
    % & \multicolumn{2}{c}{\textbf{O ($\mu$|$\sigma$)}} 
    % & \multicolumn{2}{c}{\textbf{C ($\mu$|$\sigma$)}} 
    % & \multicolumn{2}{c}{\textbf{E ($\mu$|$\sigma$)}} 
    % & \multicolumn{2}{c}{\textbf{A ($\mu$|$\sigma$)}} 
    % & \multicolumn{2}{c}{\textbf{N ($\mu$|$\sigma$)}} \\

    \cmidrule(lr){2-2} \cmidrule(lr){3-4} \cmidrule(lr){5-6} \cmidrule(lr){7-8} \cmidrule(lr){9-10} \cmidrule(lr){11-12}
    %& $\mu$ ($\sigma$) & \multicolumn{1}{c}{$\mu$ ($\sigma$)} & \multicolumn{1}{c}{$\mu$ ($\sigma$)} & \multicolumn{1}{c}{$\mu$ ($\sigma$)} & \multicolumn{1}{c}{$\mu$ ($\sigma$)} & \multicolumn{1}{c}{$\mu$ ($\sigma$)} & \multicolumn{1}{c}{$\mu$ ($\sigma$)} & \multicolumn{1}{c}{$\mu$ ($\sigma$)} & \multicolumn{1}{c}{$\mu$ ($\sigma$)} & \multicolumn{1}{c}{$\mu$ ($\sigma$)} & \multicolumn{1}{c}{$\mu$ ($\sigma$)} \\
    %\midrule

Call & 0.99 (0.18) & \cellcolor{gray}\textbf{0.97 (0.18)} & 0.87 (0.34) & \cellcolor{gray}\textbf{0.96 (0.21)} & \cellcolor{gray}\textbf{0.93 (0.25)} & 0.96 (0.22) & 0.52 (0.50) & \cellcolor{gray}\textbf{0.99 (0.11)} & 1.01 (0.15) & 0.97 (0.19) & \cellcolor{gray}\textbf{0.97 (0.17)} \\

Coffee & 0.97 (0.21) & 0.99 (0.15) & 0.91 (0.30) & 0.95 (0.21) & \cellcolor{gray}\textbf{1.07 (0.26)} & 1.00 (0.20) & 0.79 (0.42) & 1.00 (0.11) & 0.95 (0.26) & 0.99 (0.16) & 0.99 (0.11) \\

Creative & 1.04 (0.21) & \cellcolor{gray}\textbf{1.07 (0.26)} & \cellcolor{gray}\textbf{0.90 (0.32)} & \cellcolor{gray}\textbf{1.00 (0.09)} & \cellcolor{gray}\textbf{1.00 (0.18)} & \cellcolor{gray}\textbf{1.00 (0.12)} & 0.97 (0.29) & \cellcolor{gray}\textbf{1.01 (0.08)} & \cellcolor{gray}\textbf{1.00 (0.13)} & \cellcolor{gray}\textbf{1.00 (0.16)} & \cellcolor{gray}\textbf{1.00 (0.09)} \\

% Email & 1.03 (0.28) & \cellcolor{gray}\textbf{0.98 (0.17)} & \cellcolor{gray}\textbf{0.99 (0.18)} & \cellcolor{gray}\textbf{0.99 (0.16)} & \cellcolor{gray}\textbf{0.99 (0.20)} & \cellcolor{gray}\textbf{0.94 (0.25)} & \cellcolor{gray}\textbf{0.87 (0.37)} & \cellcolor{gray}\textbf{0.99 (0.13)} & 1.05 (0.25) & 1.03 (0.21) & \cellcolor{gray}\textbf{0.98 (0.13)} \\

Exercise & 0.98 (0.15) & 0.99 (0.08) & 0.83 (0.38) & 0.97 (0.17) & 0.98 (0.14) & 1.00 (0.00) & 0.60 (0.49) & 0.99 (0.13) & 0.98 (0.14) & 0.98 (0.15) & 0.99 (0.09) \\

Reading & 1.00 (0.11) & 1.01 (0.12) & 0.89 (0.32) & 0.98 (0.16) & 1.01 (0.13) & 1.01 (0.13) & \cellcolor{gray}\textbf{1.01 (0.20)} & 1.00 (0.08) & 0.98 (0.13) & 1.01 (0.15) & 1.00 (0.10) \\

Lunch & 1.01 (0.08) & 1.00 (0.04) & 1.01 (0.09) & 1.01 (0.08) & 1.00 (0.09) & 1.00 (0.04) & 1.01 (0.10) & 1.00 (0.04) & 1.00 (0.09) & 1.00 (0.06) & 1.00 (0.04) \\

Meeting & 1.00 (0.16) & \cellcolor{gray}\textbf{0.97 (0.16)} & 0.98 (0.18) & 1.00 (0.09) & 0.94 (0.27) & 1.00 (0.17) & 0.52 (0.50) & \cellcolor{gray}\textbf{0.99 (0.12)} & \cellcolor{gray}\textbf{1.04 (0.20)} & \cellcolor{gray}\textbf{0.97 (0.17)} & 0.98 (0.15) \\

Break & 1.02 (0.23) & 1.02 (0.22) & \cellcolor{gray}\textbf{0.90 (0.33)} & \cellcolor{gray}\textbf{0.98 (0.21)} & \cellcolor{gray}\textbf{1.10 (0.31)} & 1.01 (0.22) & \cellcolor{gray}\textbf{1.25 (0.48)} & 1.04 (0.23) & \cellcolor{gray}\textbf{0.94 (0.27)} & \cellcolor{gray}\textbf{1.12 (0.35)} & 1.03 (0.23) \\

Personal & 1.04 (0.26) & 1.06 (0.26) & 0.97 (0.34) & 1.05 (0.28) & 1.07 (0.34) & 1.08 (0.30) & \cellcolor{gray}\textbf{1.49 (0.68)} & 1.06 (0.26) & \cellcolor{gray}\textbf{1.00 (0.18)} & \cellcolor{gray}\textbf{1.12 (0.37)} & \cellcolor{gray}\textbf{1.11 (0.35)} \\

Plan & 1.04 (0.20) & \cellcolor{gray}\textbf{1.01 (0.11)} & \cellcolor{gray}\textbf{1.00 (0.13)} & \cellcolor{gray}\textbf{1.00 (0.09)} & \cellcolor{gray}\textbf{0.94 (0.24)} & \cellcolor{gray}\textbf{0.98 (0.16)} & \cellcolor{gray}\textbf{0.89 (0.31)} & \cellcolor{gray}\textbf{1.00 (0.13)} & \cellcolor{gray}\textbf{1.00 (0.13)} & 1.01 (0.17) & \cellcolor{gray}\textbf{1.00 (0.08)} \\

Reflect & 1.09 (0.29) & 1.05 (0.23) & \cellcolor{gray}\textbf{1.03 (0.21)} & 1.06 (0.25) & \cellcolor{gray}\textbf{0.99 (0.15)} & \cellcolor{gray}\textbf{1.02 (0.15)} & \cellcolor{gray}\textbf{1.41 (0.59)} & 1.10 (0.32) & \cellcolor{gray}\textbf{1.03 (0.18)} & \cellcolor{gray}\textbf{1.16 (0.38)} & 1.08 (0.27) \\

Research & 1.01 (0.14) & 1.03 (0.18) & 0.99 (0.12) & 1.00 (0.06) & \cellcolor{gray}\textbf{0.95 (0.22)} & 0.95 (0.22) & 1.00 (0.23) & 0.99 (0.10) & 1.00 (0.14) & 1.01 (0.18) & \cellcolor{gray}\textbf{0.99 (0.10)} \\

Media & 1.02 (0.19) & 1.00 (0.13) & \cellcolor{gray}\textbf{0.86 (0.35)} & \cellcolor{gray}\textbf{0.96 (0.19)} & \cellcolor{gray}\textbf{1.19 (0.43)} & \cellcolor{gray}\textbf{1.06 (0.26)} & \cellcolor{gray}\textbf{0.69 (0.46)} & \cellcolor{gray}\textbf{0.99 (0.13)} & 1.00 (0.14) & 1.03 (0.21) & \cellcolor{gray}\textbf{0.99 (0.09)} \\

Collab. & 1.02 (0.17) & \cellcolor{gray}\textbf{1.00 (0.08)} & \cellcolor{gray}\textbf{0.99 (0.11)} & \cellcolor{gray}\textbf{1.00 (0.06)} & \cellcolor{gray}\textbf{0.97 (0.20)} & 1.02 (0.13) & \cellcolor{gray}\textbf{0.61 (0.49)} & \cellcolor{gray}\textbf{1.00 (0.06)} & \cellcolor{gray}\textbf{1.00 (0.04)} & \cellcolor{gray}\textbf{0.99 (0.13)} & \cellcolor{gray}\textbf{1.00 (0.04)} \\

Work & 1.18 (0.46) & \cellcolor{gray}\textbf{0.89 (0.36)} & 1.18 (0.40) & \cellcolor{gray}\textbf{1.11 (0.33)} & \cellcolor{gray}\textbf{0.92 (0.36)} & \cellcolor{gray}\textbf{0.93 (0.30)} & \cellcolor{gray}\textbf{0.78 (0.43)} & \cellcolor{gray}\textbf{0.95 (0.27)} & \cellcolor{gray}\textbf{1.32 (0.52)} & \cellcolor{gray}\textbf{1.02 (0.30)} & \cellcolor{gray}\textbf{1.00 (0.18)} \\

\midrule

% \multirow{2}{*}{\textbf{Null}} & Accept & 9 & 9 & 8 & 5 & 10 & 7 & 8 & 8 & 9 & 7 \\
% & Reject & 7 & 7 & 8 & 11 & 6 & 9 & 8 & 8 & 7 & 9 \\

Reject & ... & 6 & 6 & 7 & 10 & 5 & 8 & 7 & 8 & 7 & 8 \\

    \bottomrule
  \end{tabular}
  \caption{Individual task frequency per OCEAN (+/-) condition with sample size $n = 500$. Values are mean ($\mu$) with std. dev. ($\sigma$) in parentheses. Highlighted cells in gray denote statistically significant deviations ($p$ $\leq$ 0.05) from the corresponding task frequency within the control (Neutral) condition.}
  \label{tab:chi-square_task_frequency}
  \vspace{-5pt}
\end{table*}

In each experimental group, the \textit{duration} and \textit{frequency} of at least 5 and 7 tasks significantly differed from the control, respectively. This indicates the induction of personality, based on FFM, notably affects planning-based behaviours on both of these metrics given the downstream task presented herein. Many of these differences correlated with the expected changes for the specific OCEAN trait under evaluation. For instance, positively inducing \underline{C}onscientiousness increased the average duration ($\mu$) of the \textit{Work} task (85.1m vs. 63.9m) while slightly reducing its variance ($\sigma$) (19.0 vs. 19.2). Conversely, negative induction resulted in an increased average duration ($\mu$) (66.3m vs. 63.9m) with a higher variance ($\sigma$) (23.3 vs. 19.2). Additionally, non-work tasks (\textit{e.g.}, Break, Personal Time) were scheduled for longer periods.
\section{Discussion} \label{Sec:Discussion}

This study demonstrates the controlled induction of personality traits, based on FFM, can produce distinctly different planning-based behaviours within an LLM. This is essential for the deceptive agents herein proposed, operated by the SANDMAN architecture, to be effective in their capacity to create plausibly deniable behaviours and misinformation which cannot be distinguished from human and machine. The aim hereby is to enable defenders the capability to craft and refine various simulacra personas of autonomous agents in security-focused applications. While the central focus is on deploying decoys to gather intelligence on attackers, the concept and research herein raises question toward the efficacy of low-cost, large-scale deployment of deceptive agents to achieve a dazzling effect toward adversaries. Here, numerous agents operate autonomously to simulate entire networks of interconnected systems and individuals, thereby making it difficult for attackers to distinguish between real assets and decoys.

Lastly, it must be noted that this study is observational in nature. Its central aim is to investigate whether induced personas within an LLM presents considerable effect upon planning-based behaviour within a downstream task. Exploration of any observed correlation or relationship between a given OCEAN trait and associated output is suited toward future work, outlined below.

\subsection{Future Work}

As discussed, further examination is warranted to understand how certain personality traits, and combinations thereof, modify task scheduling behaviour and whether these remain consistently aligned with expectations based on extant and emerging understandings regarding the underlying trait induced to the LLM. Additional dependent variables should be explored to characterise and evaluate the output schedule populations comprehensively. While task duration and frequency are valuable metrics, other measures are required for a more thorough comparison.

Currently, the SANDMAN decision engine processes schedules sequentially. Future work will focus on enhancing this decision-making task by incorporating LLMs to account for execution context and personality traits, leading to more complex behaviours and effectively distinguishing between intention and action within the deceptive agent. Future research will also involve implementing multi-agent communication to create a realistic simulacrum of a community exhibiting human-like behaviour. Incorporating vision-based models and other modalities will support complete autonomic behaviour and reasoning, enabling more intricate tasks and richer interactions.

Lastly, real-world deployment of SANDMAN against actual observers, such as potential adversaries within safe and sandboxed virtual environments, will provide valuable insights into the practical effectiveness and limitations of the system, particularly within a defense-oriented context predicated on denial, deceit, and misinformation. Defining and measuring the "believability" or "plausibility" of agent behaviour will be crucial for assessing how convincingly Deceptive Agents mimic human actions. Incorporating dynamic task chaining and adaptive learning capabilities will enable agents to continuously learn from previous decisions and subsequent interactions to thus adapt their behaviour, making the agents more resilient and unpredictable, further complicating attackers' efforts. Future work will thus focus on advancing SANDMAN's architecture and assessing its capabilities as a fully autonomous deceptive agent, enhancing its realism, adaptability, and effectiveness in cyber deception.

\section{Conclusion} \label{Sec:Conclusion}

This paper introduces the concept of \textit{Deceptive Agents}–a new class of autonomous agents leveraging LLMs as its central controller whose purpose is to deceive adversaries by exhibiting plausible, human-like behaviour. Agents operate on a novel architecture, inspired by the CoALA framework, which offers an extensible, modular platform for developing language agents. This study highlights the use of LLMs in generating context relevant to the operation of the deceptive agent and, importantly, utilises LLMs for task planning, which is influenced by the induction of one of the Big-Five (OCEAN) personality traits, based on FFM. The work introduces a schema for personality prompt generation that produces statistically significant schedule populations in terms of task frequency and duration. The results underscore the utility and effectiveness of using LLMs in such decision-making processes in Language Agents, employing personality traits as a control mechanism to craft distinct personas.
\clearpage
\section{Limitations} \label{Limitations}

In this work, we introduced SANDMAN, a novel architecture for developing deceptive agents designed to mimic human behaviour in digital environments. While this study extends prior research in autonomous agents, several limitations accompany the current implementation and evaluation.

\paragraph{Dependency on LLMs} SANDMAN relies heavily on LLMs for decision-making. Any imperfections in these models, such as biases or inaccuracies, can be mirrored in the agents' behaviours, potentially replicating existing stereotypes or flawed behavioural patterns, which is particularly concerning for deceptive agents.

\paragraph{Static Nature of Agent Scheduling} Our investigation focused on the initial planning process, where agents generate schedules based on induced personality traits. This static approach does not reflect the dynamic nature of human activities. Humans continuously adjust their schedules in response to new information and unforeseen events. SANDMAN agents' inability to adapt in real-time limits the realism of their actions.

\paragraph{Isolated Effect of Single-Agent Environments} SANDMAN agents currently operate independently without interacting with other agents. This isolation is a significant departure from real-world environments, particularly workplaces, where interactions and collaborations influence behaviour and task management. The lack of multi-agent interaction capabilities restricts the agents' utility in more complex scenarios.

\paragraph{Overemphasis on Personality} The assumption that personality alone dictates detailed daily schedules and actions overlooks other critical factors. Personal interests, relationships, workplace dynamics, and spontaneous decisions play significant roles in shaping human behaviour. Sole emphasis on personality may oversimplify human behaviour, leading to less realistic agent actions.

\paragraph{Evaluation and Validation Challenges} Evaluating SANDMAN agents is constrained by the simplistic scenarios in which they operate. More robust testing frameworks with actual observers are needed to assess these agents in varied environments. Additionally, the criteria for "believable" or "plausible" behaviour by a language agent in a digital environment need to be rigorously defined and measured.
\section{Ethics}

The design of autonomous agents, specifically "Deceptive Agents" as outlined in our SANDMAN architecture, offers significant capabilities for enhancing cyber defense through strategic deception. However, due to the human-like nature of these agents, a thorough examination of the ethical implications and societal impact is necessary.

\paragraph{\textbf{Ethical Use of Deception}}

Deceptive Agents are designed to deceive unauthorised users attempting to access or compromise digital systems, extending existing deception technologies like honeypots \cite{spitzner2003honeypots}. The primary purpose of these agents is defensive, not malicious. They mimic human behaviour to create plausible yet non-functional digital decoys, misleading attackers to protect sensitive data and systems. This approach is ethically justified on the principle of "rightful deception" in response to unauthorised and malicious actions, where the deceived party has no legitimate claim to truth due to their unethical intent.

\paragraph{\textbf{Ethical Use of SANDMAN}}

SANDMAN agents are designed to operate in isolated environments, strictly for deceiving malicious actors. Although the architecture is general-purpose and modifiable, it is not intended for use as a "virtual employee" in real networks. Using a Gen-AI agent as an actual employee raises ethical concerns about accountability and responsibility, which should be avoided until further research on the feasibility of Gen-AI in the workplace is conducted.

\paragraph{\textbf{Exacerbated Misinformation Generation}}

There is a risk that Deceptive Agents could exacerbate existing risks associated with Gen-AI, such as deepfakes, misinformation generation, and tailored persuasion \cite{park2023generative}.

\paragraph{\textbf{Controlled Behaviour}}

There is a risk of Deceptive Agents operating outside their intended scope or generating concerning material due to their interaction with digital environments. If entirely driven by LLMs, safety constraints are applied to minimise this risk.

\clearpage
% \bibliography{refs}
% \bibliographystyle{acl_natbib}

\end{document}